\documentclass{article}

    \PassOptionsToPackage{numbers, compress}{natbib}

    \usepackage[preprint]{neurips_2024}

\usepackage[utf8]{inputenc} 
\usepackage[T1]{fontenc}    
\usepackage{hyperref}       
\usepackage{url}            
\usepackage{booktabs}       
\usepackage{amsfonts}       
\usepackage{nicefrac}       
\usepackage{microtype}      
\usepackage{xcolor}         
\usepackage{multirow}
\usepackage{tablefootnote}
\usepackage{footnote}
\usepackage{enumitem}
\usepackage{graphicx}
\usepackage{bm}
\usepackage{wrapfig}

\setlist[itemize]{leftmargin=*}

\title{\Large{AIOS Compiler: LLM as Interpreter for Natural Language Programming and Flow Programming of AI Agents}}

\author{%
  Shuyuan Xu \thanks{Both authors contributed equally to this work.} \\
  Rutgers University \\
  \texttt{shuyuan.xu@rutgers.edu} \\
  \And 
  Zelong Li \footnotemark[1]\\
  Rutgers University \\
  \texttt{zelong.li@rutgers.edu} \\
  \AND 
  Kai Mei \\
  Rutgers University \\
  \texttt{kai.mei@rutgers.edu} \\
  \And
  Yongfeng Zhang \thanks{Corresponding author} \\
  Rutgers University \\
  \texttt{yongfeng.zhang@rutgers.edu}
}

\begin{document}

\maketitle

\begin{abstract}
  

  
  Since their inception, programming languages have trended towards greater readability and lower barriers for programmers. Following this trend, natural language can be a promising type of programming language that provides great flexibility and usability and helps towards the democracy of programming.
  However, the inherent vagueness, ambiguity, and verbosity of natural language pose significant challenges in developing an interpreter that can accurately understand the programming logic and execute instructions written in natural language. 
  Fortunately, recent advancements in Large Language Models (LLMs) have demonstrated remarkable proficiency in interpreting complex natural language. 
  Inspired by this, we develop a novel system for \textbf{Co}de \textbf{R}epresentation and \textbf{E}xecution (\textbf{CoRE}), which employs LLM as interpreter to interpret and execute natural language programs (NLPg). 
  The proposed system unifies natural language programming, pseudo-code programming, and flow programming under the same representation for constructing language agents, 
  while LLM serves as the interpreter to interpret and execute the agent programs.
  In this paper, we begin with defining the programming syntax that structures natural language instructions logically. During the execution, we incorporate external memory to minimize redundancy. Furthermore, we equip the designed interpreter with the capability to invoke external tools, compensating for the limitations of LLM in specialized domains or when accessing real-time information.
  This work is open-source at \url{https://github.com/agiresearch/CoRE}, \url{https://github.com/agiresearch/OpenAGI}, and \url{https://github.com/agiresearch/AIOS}.

\end{abstract}

\section{Introduction}

\begin{figure}[t]
    \centering
    \includegraphics[width=\textwidth]{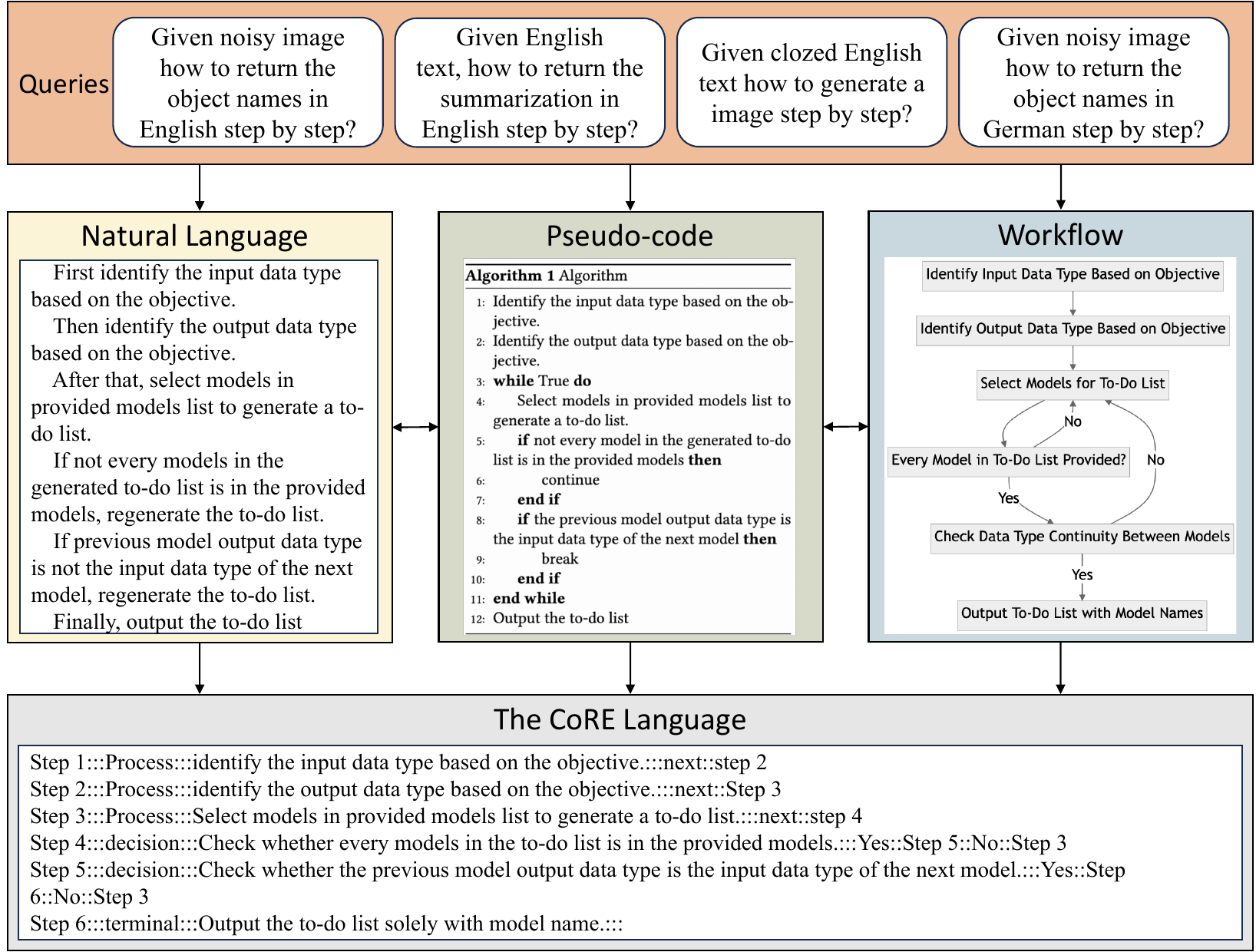}
    \caption{In our CoRE system, we design the CoRE language to unify natural language programming, pseudo-code programming, and flow programming in the same syntax representative. We use the program for OpenAGI \citep{openagi} platform as an example.}
    \vspace{-10pt}
    \label{fig:NLPg}
\end{figure}

Programming is crucial for computers as it enables them to execute specific tasks based on a predefined set of instructions. It allows us to utilize logical algorithms to enable computers to solve problems. Programming has evolved significantly since its inception, with new technologies and innovations driving its growth. Initially, programming languages were based on binary machine language, such as punched cards, which can be directly executed by the machine. However, 
machine language was hardly readable to humans. Subsequently, low-level programming languages, such as assembly language, use mnemonic instructions and operands to represent machine code, which enhances the readability \citep{hennessy2011computer}. However, due to the requirement of controlling memory locations and registers, assembly language still has a high entry barrier for programmers.
With the design of high-level programming languages like C/C++, Java and Python, coding has become more user-friendly and efficient. They offer programmers a more productive and accessible approach, leading to increased participation in programming and software development. Consequently, programming languages are becoming more integrated into everyday life.

From the history of programming languages, we can observe a clear trend toward increased usability, readability, and democracy of programming. Following this trend, natural language can be a desirable choice for coding due to its accessibility, readability, and minimal training requirements for programmers. However, the application of natural language programming presents challenges due to the inherent vagueness, ambiguity, and verbosity of natural language. The recently emerged Large Language Models (LLMs) serve as a solution to this challenge due to their extraordinary capability in language understanding \cite{ouyang2022training, chung2024scaling}, tool use and function calling \cite{openagi,qin2023toolllm},
as well as interacting with human or environments \cite{ross2023programmer, driess2023palm}. 
In this work, we propose a novel system for Code Representation and Execution (CoRE), which takes LLM as the interpreter to interpret and execute the instructions in natural language, enabling agent programming in natural language.

CoRE can be used for natural language programming, pseudo-code programming, and flow programming, as the three forms of agent programs unify into our CoRE language, as shown by the example in Figure \ref{fig:NLPg}. 
In the realm of programming, the fundamental task involves designing and developing logically structured instructions to address specific problems. 
Natural language programming offers a method where instructions are formulated in everyday language, making the code intuitive and accessible. When we structure all natural language instructions in a logical way, it inherently mirrors the essence of pseudo-code programming. Pseudo-code, by design, simplifies the coding process by stripping down syntax complexities and focusing on the algorithmic logic for easy understanding. Therefore, when the instructions are expressed in natural language, the structured instructions can be identified as pseudo-code. Moreover, pseudo-code shares a direct relationship with flow programming, as it essentially represents the algorithm's logic that can seamlessly be visualized as a workflow. Workflow, in turn, provides a graphical representation of the step-by-step execution of programs, emphasizing the decision-making visualization process and the flow of control across the program. 

\begin{figure}[t]
    \centering
    \includegraphics[width=\textwidth]{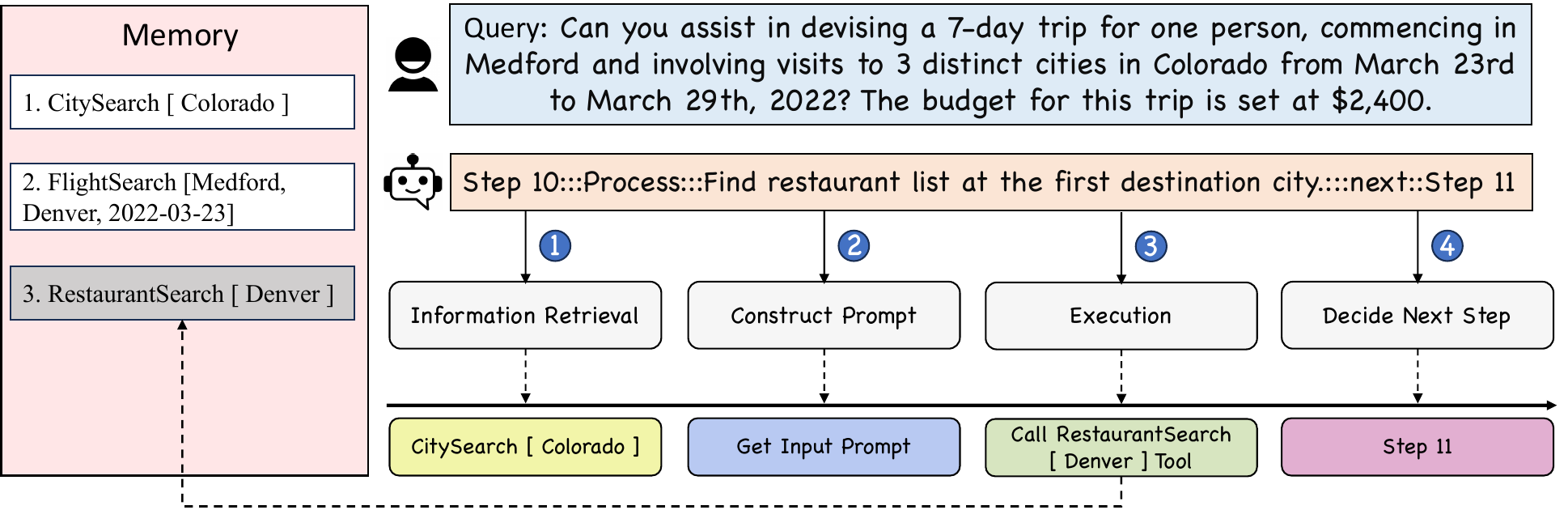}
    \caption{An example showing how the CoRE system executes one step.}
    \label{fig:OneStepExample}
    \vspace{-10pt}
\end{figure}

We face several significant challenges when designing the novel system for natural language programming with LLM as an interpreter. First, how to represent the logic of the program using natural language instructions.
To tackle this issue, we design a set of programming syntax to logically structure natural language instructions, and unify the natural language programming, pseudo-code programming, and flow programming in the same representation. 
Second, given that the programs consist of step-by-step instructions, it is crucial to make sure that each step is executed according to its corresponding instruction. 
To ensure precise execution of the instructions in natural language for each step, we design two additional components: one for retrieving information from memory, and the other for invoking external tools. Considering the LLM's limitation on the number of input tokens (context window size), including all runtime information in the input prompt is impractical. To address this problem, we store a large volume of intermediate results in temporary memory, retrieving relevant information as needed in subsequent steps \cite{wang2023augmenting, mei2024aios, lewis2020retrieval, borgeaud2022improving}. Besides, while LLMs excel at processing textual information, they often fall short in tasks that require domain-specific knowledge or up-to-date information \cite{ge2023llm}. To mitigate these limitations, we enable the LLM to utilize external tools to solve the problems \cite{openagi, qin2023toolllm, liang2023taskmatrix, ahn2022can}. 
Finally, when executing the natural language program, incorrectly determining the next step can lead to different final results. We solve this problem by demanding the LLM interpreter to evaluate the current results so as to identify the most suitable subsequent step.
The overall execution pipeline of CoRE is depicted in Figure \ref{fig:OneStepExample}. 

In summary, the key contributions of the work are listed as follows:

\begin{itemize}[leftmargin=*]
    \item We design a CoRE language that unifies natural language programming, pseudo-code programming and flow programming. The CoRE language logically structures natural language instructions. 
    \item We propose the CoRE system, which utilizes Large Language Model (LLM) as an interpreter to interpret and execute instructions step-by-step. During execution, the LLM follows the instructions and leverages both information retrieval and external tools to enhance its effectiveness.
    \item We verify the effectiveness and efficiency of our system based on public benchmark datasets. Specifically, we employ our proposed system for agent task solving based on natural language programs, showcasing its practical capabilities.
\end{itemize}

In the following part of this paper, we first provide the related work in Section \ref{sec:related_work}. In Section \ref{sec:core} we present the CoRE framework and how the framework can be applied to LLM agents. We provide the experimental results in Section \ref{sec:experiment}, and conclude the work together with future directions in Section \ref{sec:conclusions}.



\section{Related Work}
\label{sec:related_work}

\subsection{Natural Language Programming}
Research in natural language programming \cite{heidorn1976automatic, bruckman1999should, vadas2005programming, li2015nlp, desai2016program, ernst2017natural} primarily focus on addressing the ambiguity in translating natural language into programming language statements.
\citeauthor{heidorn1976automatic} \cite{heidorn1976automatic} proposes to adopt heuristic NLP encoding and decoding rules to develop an automatic programming system that can accept natural language dialogues. \citeauthor{vadas2005programming} \cite{vadas2005programming} introduce a prototype system that can translate certain English instructions into
executable Python code using Combinatory Categorial Grammar (CCG) parser, which uses unrestricted syntax to cover a wide range of user instruction semantics. 
\citeauthor{mihalcea2006nlp} \cite{mihalcea2006nlp} implement a procedural natural language programming system to convert natural language to programming language. 
Early natural language programming techniques are restricted in extensibility by the need to create domain-specific languages (DSLs). To avoid the problems of repeatedly designing new DSLs, \citeauthor{desai2016program} \cite{desai2016program} propose a general generative framework for constructing a program that takes natural language input and produces the expressions in the target DSL. Further, \citeauthor{ernst2017natural} \cite{ernst2017natural} leverages neural networks, i.e., the recurrent neural networks (RNN), to convert English specifications of file system operations into corresponding bash commands. 

\subsection{Large Language Models and AI Agents for Problem Solving}

Large Language Models (LLMs) have emerged as powerful tools for problem solving, encompassing tasks in reasoning, planning, and code generation. LLM reasoning typically involves decomposing a complex task into a sequence of steps, also known as a reasoning chain \cite{wei2022chain}. Prominent approaches in LLM reasoning include Chain-of-Thought (CoT) and its derivatives \cite{wei2022chain, kojima2022large}. To further improve the reasoning ability of LLM, several work has been proposed. The Self-consistency method \cite{wang2022self} samples multiple reasoning paths and selects the most consistent outcome by voting. Additionally, classical data structures like trees and graphs are utilized to enhance reasoning efficiency and accuracy in fewer steps \cite{yao2024tree, besta2024graph}. Apart from reasoning, planning is also an important task that can be used to solve problems. LLM Planning involves generating a series of actions to achieve the predefined goals \cite{hao2023reasoning}. Recent advancements include direct prompting of LLMs for planning tasks, showing promising results \cite{huang2022inner, singh2023progprompt, ding2023task}. Finite state machines have been integrated into LLM to enhance the planning ability \cite{li2024formalllm, wu2024stateflow}. ReAct \cite{yao2022react} proposes to leverage external tools like search engine to enhance the LLM planning. Besides, considering the powerful ability of LLM in programming, recent work propose to generate programming code to solve problems \cite{lyu2023faithful, jojic2023gpt, liu2023llm, chen2022program, josifoski2023flows, poesia2022synchromesh, nijkamp2022codegen}. Furthermore, the ``self-reflection'' mechanism \cite{madaan2024self, paul2023refiner, shinn2023reflexion} enables LLMs to critique their own outputs, significantly enhancing performance in tasks such as reasoning \cite{besta2024graph} and code generation \cite{chen2023teaching}. In contrast to existing methods that directly use LLMs for generating solutions, the proposed CoRE system utilizes LLMs as interpreters, executing solutions designed by humans to address complex questions. This approach leverages human creativity in solution design, coupled with LLM's ability, to enhance problem-solving capabilities in natural language programming contexts.




\begin{figure}[t]
    \centering
    \includegraphics[width=.8\textwidth]{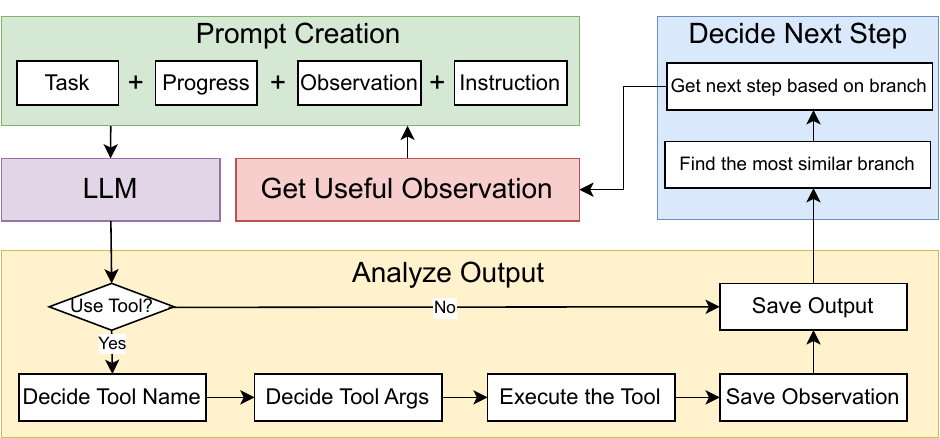}
    \vspace{-5pt}
    \caption{An overview of the CoRE LLM interpreter system.}
    \vspace{-10pt}
    \label{fig:overview}
\end{figure}

\section{The CoRE System}

\label{sec:core}
In this section, we will introduce how we define the natural language programming syntax and how to use LLM as an interpreter to interpret and execute natural language programs.

\subsection{CoRE Language Syntax}
To organize natural language instructions, we define the basic structural representation for each step, which consists of four components. An example can be found in Figure \ref{fig:NLPg}.
\begin{itemize}[leftmargin=*]
    \item \textbf{Step Name}: Each step in the program is uniquely identified by a step name. This identifier 
    is analogous to function identifiers in traditional programming languages, which facilitates navigation and reference within the program structure, ensuring that each operation within the program can be distinctly addressed and accessed.
    \item \textbf{Step Type}: The step type categorizes the nature of the operation being performed in each step, analogous to control structures in conventional programming. We define three primary step types: 
    \begin{itemize}
        \item Process: Akin to a procedural statement in traditional programming, this step type executes a specific operation and transitions to the next specified step.
        \item Decision: Corresponding to conditional statements (e.g., ``if-else''), this step involves branching the program flow based on evaluated conditions, leading to multiple potential paths.
        \item Terminal: Similar to the ``end'' or ``return'' statement, this step marks the conclusion of the program, indicating that no further steps are to be executed.
    \end{itemize}
    \item \textbf{Step Instruction}: The step instruction explicates the task to be conducted at a step. This component is integral as it provides the instruction and content for execution, paralleling the statement block in traditional programming languages. By demonstrating operations in natural language, NLPg lowers the barrier to programming, making it more readable for non-expert programmers.
    \item \textbf{Step Connection}: Step connections define the progression from one step to another, establishing the flow of the program execution. In process steps, a single subsequent step is specified. In decision steps, multiple pathways are delineated based on conditions. Terminal steps, by definition, do not lead to any future steps, indicating the end of program execution.
\end{itemize}

For each step in the program, the above four components are separated by ``:::'' (as illustrated in the CoRE language in Figure \ref{fig:NLPg}). Other special tokens can also be used to separate different components.

\begin{figure}[t]
    \centering
    \includegraphics[width=0.9\textwidth]{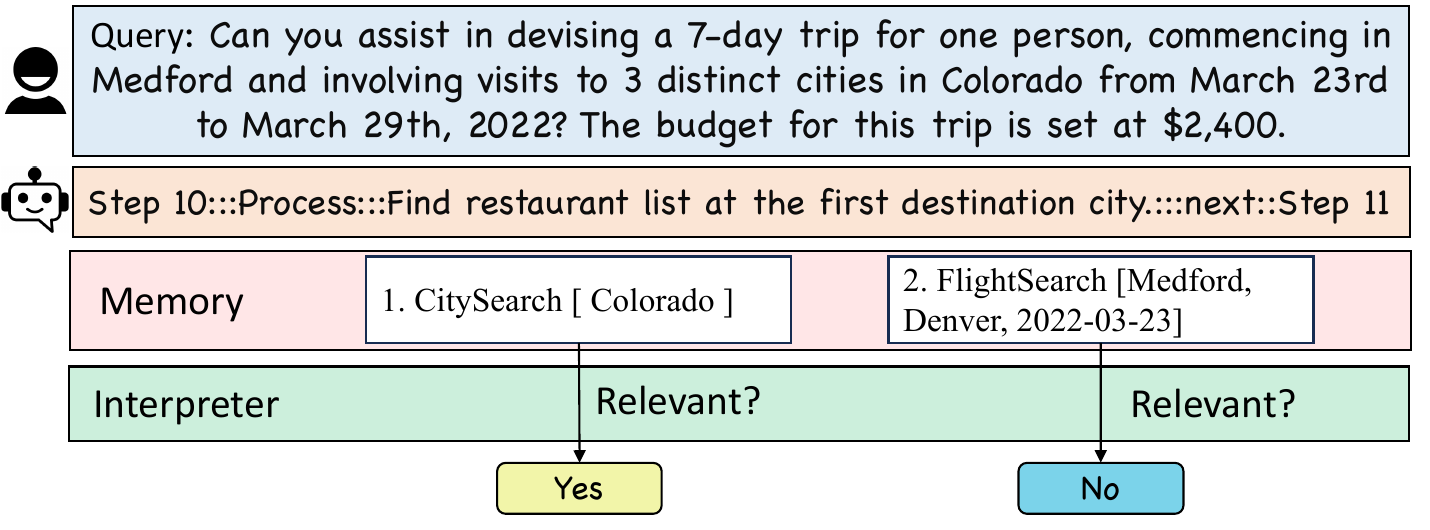}
    \vspace{-5pt}
    \caption{An example showing how the CoRE system retrieves relevant information.}
    \label{fig:observation}
\end{figure}

In programming languages, there are three basic control constructs in programming \cite{dahl1972structured, prather1997regular}: sequence, selection and iteration. These three basic constructs can be easily designed within the CoRE language. 
\begin{itemize}
    \item \textbf{Sequence}: Sequence in programming is the execution of statements in a linear order, with each statement leading to the next. In the CoRE framework, this construct is designed by setting the ``Step Connection'' to point to the subsequent step. Each step operates under the \textit{Process} type until the sequence concludes.

    \item \textbf{Selection}: Selection in programming languages facilitates conditional branching, allowing the program to execute different sequences of steps based on specific conditions. This is implemented using the \textit{Decision} step type where the ``Step Connection'' part explicitly outlines multiple potential paths. Each branch is defined by a condition stated within the ``Step Connection'' part, guiding the program flow to various steps depending on the conditions. 

    \item \textbf{Iteration}: Iteration involves repeating a set of operations until a certain condition is met, akin to loops in conventional programming. In the CoRE framework, we utilize a step with the \textit{Decision} type to assess whether the loop condition has been fulfilled. At the end of one loop cycle, the ``Step Connection'' is configured to point back to the previous \textit{Decision} step, thereby enabling the continuation of the loop.
\end{itemize}

\subsection{LLM as Interpreter}

In this section, we will discuss how the CoRE system utilizes a Large Language Model (LLM) as an interpreter to execute programs written in the CoRE language. We will demonstrate the execution of a single step within the CoRE system, which is illustrated in Figure \ref{fig:overview}. More specifically, the system executes a single step in four procedures. First of all, the interpreter determines the useful information to execute the current step. Then the interpreter will integrate all relevant information to construct the prompt. Based on the generated prompt, the interpreter will generate response and may utilize tools to execute the current step. Finally, after executing the current step, the interpreter will determine the next step based on step type and execution results. We will introduce the four parts in details.

\subsubsection{Observation Retrieval from Memory}
    

This initial procedure is critical since it sets the stage for the entire execution process of the current step. Figure \ref{fig:observation} shows an example. The system's memory serves as a repository of all prior observations related to the program, where the observation represents the results of tool execution, such as search results. During this phase, the interpreter scans the memory to identify records that are relevant to the current instruction. This selective retrieval ensures that the interpreter's decisions are informed by accurate and contextually relevant data, which is crucial for the successful execution of the program.

\begin{figure}[t]
    \centering
    \includegraphics[width=0.9\textwidth]{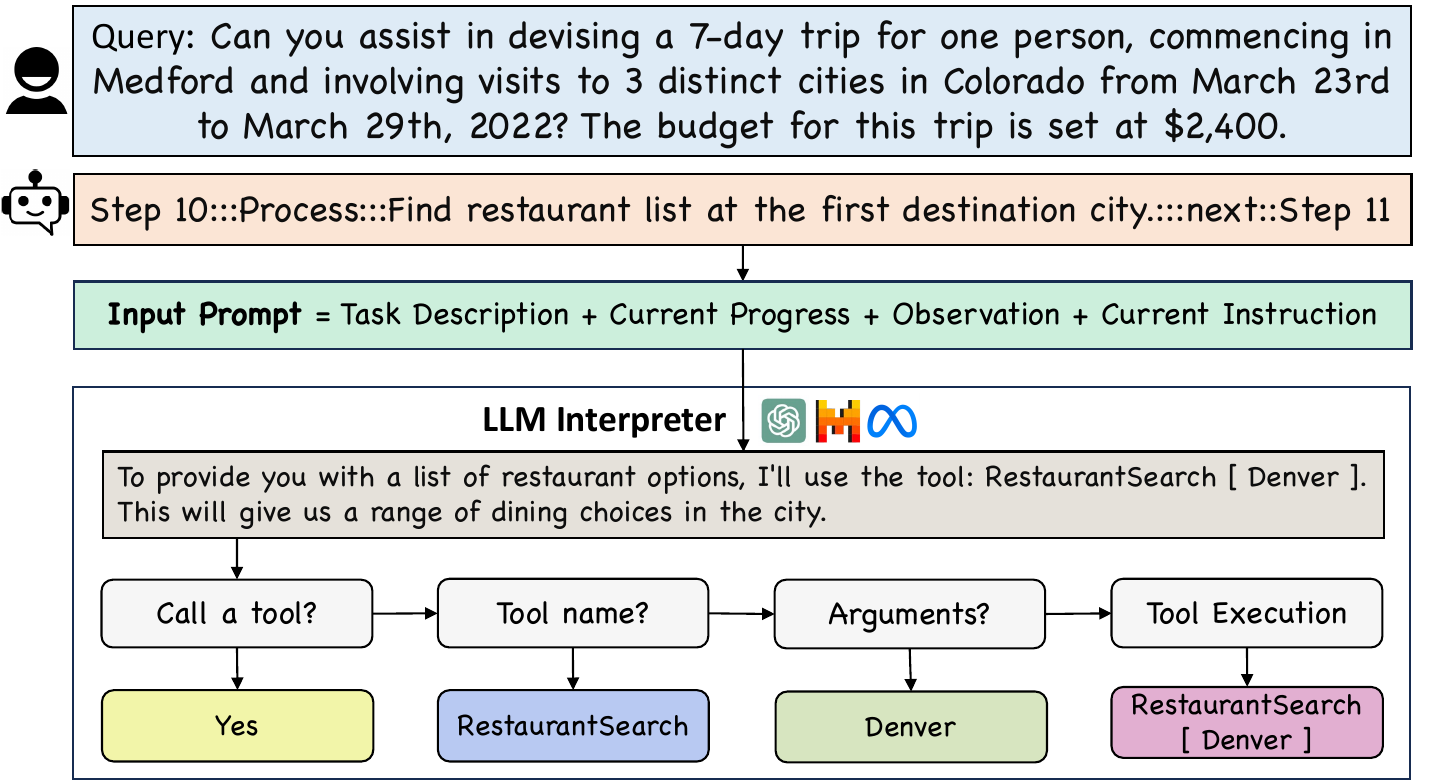}
    \caption{An example showing how the CoRE system analyze the output from the LLM interpreter.}
    \label{fig:execution}
\end{figure}

\subsubsection{Input Prompt Construction}
Constructing the prompt is essentially about synthesizing the information into a comprehensive and coherent query that the LLM can understand and respond to effectively. This involves combining multiple information into a single, structured prompt that guides the LLM towards generating the most appropriate and contextually relevant response. In the CoRE system, the interpreter constructs a detailed prompt with four elements:
\begin{itemize}
    \item \textbf{Task Description}: The query that defines the entire program, acting as the primary input to guide the system's operations.
    \item \textbf{Current Progress}: Summarizes the previous steps including what has been done or decided, helping maintain a narrative flow.
    \item \textbf{Observation}: This part may not be included in every step. When relevant information is retrieved from the memory by the interpreter, it is incorporated here.
    \item \textbf{Current Instruction}: Specifies the action to be taken in natural language, directing the interpreter on how to proceed in the current step.
\end{itemize}

\subsubsection{Output Analysis}

While the LLM can generate direct responses, complex tasks may require capabilities beyond its immediate scope. Incorporating the use of specialized tools when necessary extends the LLM's capabilities, allowing the system to handle a broader range of tasks effectively. A demonstrative example of the execution process is shown in Figure \ref{fig:execution}. In the CoRE system, the interpreter will make a decision about if or not to employ specialized tools based on the LLM's initial response and the demands of the task at hand, which ensures that the system remains highly functional and versatile, actively solving problems rather than merely processing the language prompt for the current step. Specifically, if tool usage is warranted, the system will select the suitable tool, configure it with the necessary parameters, execute it, and integrate the output into the ongoing process.

\subsubsection{Branching Analysis}

\begin{figure}[t]
    \centering
    \includegraphics[width=0.8\columnwidth]{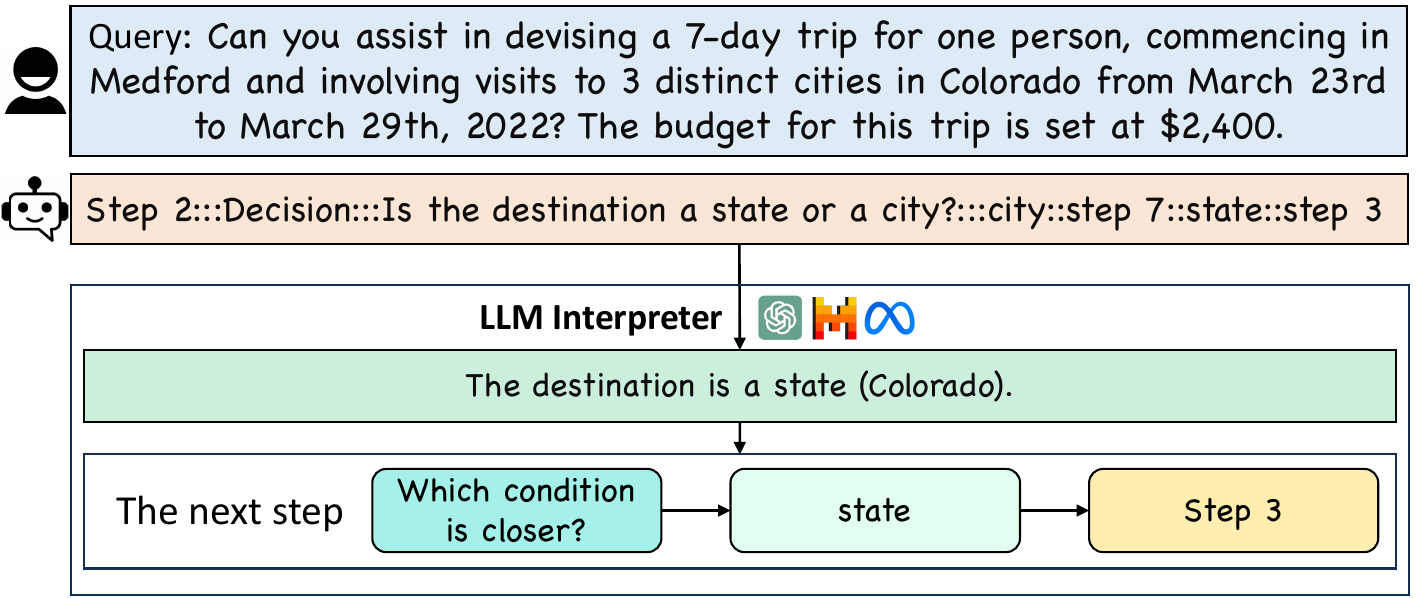}
    \vspace{-5pt}
    \caption{An example showing how the CoRE system determines the next step in the flow.}
    \label{fig:nextstep}
\end{figure}

Determining the appropriate next step in the program is critical, especially in multi-branch scenarios where different outcomes can lead to different subsequent actions. Figure \ref{fig:nextstep} shows an example. In the CoRE language interpreter, the \textit{Decision} steps indicate multiple branches with the corresponding conditions. The interpreter uses LLM to decide if the prompt satisfies the natural language described branching condition or not and which next step to take.
This adaptive approach allows the system to navigate through decision points effectively, ensuring logical progression toward the program's goals. 

\section{Experiments}
\label{sec:experiment}

\subsection{Backbone Large Language Model (LLM)}
\label{sec:backbone}

We conduct experiments on both closed-source and open-source LLMs:

\begin{itemize}[itemsep=0pt, topsep=0pt]
    \item \textbf{GPT-4} \citep{openai2023gpt4}  (Closed-source) is a generative pre-trained transformer of OpenAI
    . In this work, we use the GPT-4-1106-preview version.
    \item \textbf{Mixtral-8x7B} \citep{jiang2024mixtral} (Open-source) is a pre-trained generative Sparse Mixture of Experts with 46.7 billion parameters. 
\end{itemize}

\subsection{Planning Schema of LLMs}
\label{sec:schema}

We adopt the following LLM-based agent planning schema:

\begin{itemize}[itemsep=0pt, topsep=0pt]
    \item \textbf{Zero-shot Learning (Zero)} directly inputs the query to the LLM.
    \item \textbf{Chain-of-Thought (CoT)} \cite{wei2022chain} induces the LLM to generate a coherent language sequence that serves as a meaningful intermediate step bridging the input query and the output answer.
    \item \textbf{Few-shot Learning (Few)} presents a set of high-quality demonstrations in the prompt, each consisting of both input and desired output on the target task. 
    \item \textbf{CoRE} is our natural language programming method with LLM as an interpreter.
\end{itemize}

\subsection{Benchmark Datasets}

We conduct experiments on a benchmark dataset, \textbf{OpenAGI} \citep{openagi}. 
The OpenAGI benchmark tasks are categorized based on their output type and ground-truth label type (\textbf{Task 1}, \textbf{2}, and \textbf{3}). Then, based on different task types, different metrics are employed to gauge the performance: \textbf{CLIP Score} \citep{hessel2021clipscore}, assessing the similarity between text and image, is utilized for Text-to-Image tasks; \textbf{BERT Score} \citep{bert-score}, evaluating text generation with BERT, is applied when both data labels and the expected outputs are texts; and \textbf{ViT Score} \citep{wu2020visual} gauges the similarity between the image label and image output. 

\subsection{Implementation Details}

Our framework and all baselines are implemented by PyTorch, an open-source library. We follow the implementation setting of the OpenAGI platform \citep{openagi} for Zero-shot and few-shot learnings. We leverage the DSPy framework \cite{khattab2022demonstrate, khattab2023dspy} to apply the CoT strategy to the OpenAGI platform. We also tried Program-of-Thought \cite{chen2022program} and ReAct \cite{yao2023react} strategies on the OpenAGI platform. However, the ReAct strategy requires text observation, which is unsuitable for our OpenAGI task since some observations are in image format, and Program-of-Thought cannot generate executable codes. Thus, we did not include them as the baselines.

\subsection{Experimental Analysis}
\label{sec:exp_analysis}

The experiment results on the OpenAGI benchmark are shown in Table \ref{Table:openagi}. Each row stands for a type of task, each column represents the planning schema of an LLM interpreter, and every four columns are the results of the same LLM interpreter. From the results, we can see that our CoRE planning schema is better on average performance than any baseline under both Mixtral and GPT-4 as the interpreters. When using Mixtral as the interpreter, CoRE outperforms Zero-shot and CoT under each type of task, and is better than Few-shot learning on Task 2 and average score, though worse on Task 3 and slightly worse on Task 1. When using GPT-4 as the interpreter, CoT, Few-shot has similar performance on Task 1 and Task 3, while on Task 2 and average score, CoRE is still the best. It may be worth noting that it is unfair to compare CoRE with Few-shot learning since we do not directly provide the output format and output example in the prompt. However, even without using such examples, the CoRE planning strategy is still better than the Few-shot strategy on average. We also find that even for the same CoRE program, the system may perform differently when using different LLM as interpreters, which means that the performance of natural language programming depends on the natural language understanding ability of the LLM interpreter.

\begin{table}[t]
\small
    \centering
    \setlength{\tabcolsep}{3pt}
    \begin{tabular}{c|cccc|cccc}
    \hline
      \multirow{2.5}{*}{Metrics / Task} & \multicolumn{4}{c|}{Mixtral (open source) as LLM interpreter} &\multicolumn{4}{c}{GPT-4 (closed-source) as LLM interpreter} \\
      \cline{2-5} \cline{6-9}
      & Zero & CoT & Few & CoRE (Ours) & Zero & CoT & Few & CoRE (Ours) \\
      \hline
      Task 1 (CLIP Score) & 0.0 & 0.0 & \bm{$0.1839$} & 0.1825 & 0.0 & 0.2732 & 0.1837 & \bm{$0.3030$} \\
      Task 2 (BERT Score) & 0.1092 & 0.1987 & 0.0687 & \bm{$0.2593$} & 0.2076 & 0.2266 & 0.5277 & \bm{$0.5756$} \\
      Task 3 (ViT Score) & 0.1949 & 0.1562 & \bm{$0.5501$} & 0.2437 & 0.5058 & 0.6736 & \bm{$0.6916$} & 0.6611 \\
      Average over tasks & 0.1206 & 0.1736 & 0.1887 & \bm{$0.2483$} & 0.2378 & 0.3359 & 0.5391 & \bm{$0.5744$} \\
      \hline
      \% of Valid Plans & 23.08 & 38.46 & 46.15 & \bm{$56.92$} & 53.85 & 60.00 & 83.08 & \bm{$92.31$} \\
      \hline
    \end{tabular}
    \caption{OpenAGI \cite{openagi} benchmark task performances under different settings. Zero is for Zero-shot Learning, Few is for Few-shot Learning. The boldface numbers denote the highest score under each task type using the same LLM.}
    \label{Table:openagi}
    \vspace{-15pt}
\end{table}


\section{Conclusions and Future Work}
\label{sec:conclusions}

In this study, we introduce a novel system, CoRE, for Code Representation and Execution. CoRE is designed to bridge natural language programming, pseudo-code, and flow programming through the development of a unified CoRE language for the construction of AI Agents. CoRE leverages natural language as the programming interface, which lowers the programming barrier and advocates the democracy of programming, so that even ordinary users can create their AI Agents. Our system leverages Large Language Models (LLMs) as interpreters to process and execute natural language instructions. Throughout execution, the interpreter dynamically retrieves necessary information, utilizes appropriate external tools, and navigates through instructions based on previous outputs. The experimental outcomes validate the efficacy of the CoRE system in natural language programming.

While CoRE demonstrates promising results, it currently relies on manually crafted programs, which may introduce inefficiencies due to the inherent ambiguities of natural language. To address this, future research could explore the development of automated systems for generating natural language programming instructions. This automation would help standardize instruction clarity and precision, potentially improving system performance. Additionally, a future direction is to expand CoRE's language support to facilitate international use and implement real-time debugging features to aid in education and assist novice programmers, further broadening the system's utility and accessibility.

\bibliographystyle{ACM-Reference-Format}
\bibliography{reference}

\newpage
\appendix

\end{document}